\documentclass[conference]{IEEEtran}
\IEEEoverridecommandlockouts
\usepackage{cite}
\usepackage{amsmath,amssymb,amsfonts}
\usepackage{algorithmic}
\usepackage{graphicx}
\usepackage{textcomp}
\usepackage{xcolor}
\usepackage{multirow}
\usepackage{booktabs}       
\usepackage{caption}
\usepackage{multicol}
\usepackage{float}
\usepackage{ulem}
\usepackage[]{fancyhdr}

\def\BibTeX{{\rm B\kern-.05em{\sc i\kern-.025em b}\kern-.08em
    T\kern-.1667em\lower.7ex\hbox{E}\kern-.125emX}}

\begin{document}

\pagestyle{fancy}
\fancyhead[CO,RE]{Published as a conference paper at 2024 IEEE International Symposium On Circuits and Systems}

\title{Efficient Neural Compression with Inference-time Decoding\\
\thanks{}
}

\author{\IEEEauthorblockN{1\textsuperscript{st} Clement Metz}
\IEEEauthorblockA{\textit{CEA List, Université Paris-Saclay} \\
Palaiseau, France \\
clement.metz@univ-grenoble-alpes.fr}
\and
\IEEEauthorblockN{2\textsuperscript{nd} Olivier Bichler}
\IEEEauthorblockA{\textit{CEA List} \\
Palaiseau, France \\
olivier.bichler@cea.fr}
\and
\IEEEauthorblockN{3\textsuperscript{rd} Antoine Dupret}
\IEEEauthorblockA{\textit{CEA Leti} \\
Palaiseau, France \\
antoine.dupret@cea.fr}
}

\maketitle

\begin{abstract}
This paper explores the combination of neural network quantization and entropy coding for memory footprint minimization. Edge deployment of quantized models is hampered by the harsh Pareto frontier of the accuracy-to-bitwidth tradeoff, causing dramatic accuracy loss below a certain bitwidth. This accuracy loss can be alleviated thanks to mixed precision quantization, allowing for more flexible bitwidth allocation. However, standard mixed precision benefits remain limited due to the 1-bit frontier, that forces each parameter to be encoded on at least 1 bit of data. This paper introduces an approach that combines mixed precision, zero-point quantization and entropy coding to push the compression boundary of Resnets beyond the 1-bit frontier with an accuracy drop below 1\% on the ImageNet benchmark. From an implementation standpoint, a compact decoder architecture features reduced latency, thus allowing for inference-compatible decoding.
\end{abstract}


\section{Introduction}

Low latency applications and data privacy requirements fuel increasing demand for on-device deep networks. In turn, it raises numerous challenges: either the size or the power of the chip might be limited, or its capacity to access its memory fast enough to perform inference without significant latency degradation. The weights of a network need to be stored at all times on the edge, ready to be accessed, even though they are accessed just once during inference. Therefore, reducing their memory footprint is of utmost importance: it allows to minimize the size of the chip and thus the latency of memory access. Pruning \cite{blalock} is one way of proceeding, by removing parameters and entire neurons. Quantization \cite{krishnamoorthi2018quantizing} is another, by lowering the number of bits used to encode each parameter.
\hfill\break\indent
Neural network quantization has been thoroughly studied in the litterature. The most efficient quantization methods perform so-called ``Quantization-Aware Training'' (QAT), meaning they train the network by fake-quantizing weights and/or activations during the forward pass. Most quantization methods are uniform (constant quantization step) \cite{jin2019efficient,esser2020learned}, allowing for integer computations. Others proceed by power-of-two quantization \cite{apot}, or block quantization \cite{stock2020bit}.
\hfill\break\indent
More adaptive than rigid constant precision methods, mixed-precision quantization is a subclass of methods that extend the search space of the quantization process to the choice of bit precisions (either per layer, per channel, per tensor, etc.). Precisions are either determined by a metric that assigns an importance factor to each parameter group \cite{dong2019hawq,dong2019hawqv2}, or trained along network parameters \cite{habi2020hmq,fracbits}.
\hfill\break\indent
Further lossless compression can be achieved through source coding (also referred as entropy coding). In this case, the dataflow from the Dynamic Random Access Memory (DRAM) is decompressed just before computation. Source coding is less used at inference time because of its usually inefficient benefits-to-complexity ratio. Nonetheless, source coding for neural networks has been studied for the purpose of storing and transferring models. Deep Compression \cite{han_deep_2016} has influenced the following state of the art in this matter. Several methods use source coding at some point in their compression pipeline: \cite{oktay2020scalable} train their own reparametrization and follow it up with arithmetic coding, \cite{deepcabac} devise a full compression format, including arithmetic and Golomb coding, whereas \cite{hemp} train a neural network with entropy regularization to compress it with LZMA \cite{lzma}. 
\hfill\break\indent
In this paper, we show that source coding can be most beneficial at inference time when combined with an appropriate quantization method, and can lead to substantial compression improvement with limited hardware and latency overhead. In section \ref{methods}, a source coding scheme compatible with inference is proposed, the advantages of zero-point quantization are discussed and it is shown how to train the network relatively to an entropy objective. Our approach is underpinned by experiments on the ImageNet classification task. Sub-binary compression rates are achieved for feature extractor weights of a standard Resnet-50 with just $0.5\%$ accuracy loss on the full precision baseline.

\section{Methods} \label{methods}

\subsection{Entropy bound and ANS decoder}

\begin{figure}[h]

\begin{center}
\includegraphics[scale=0.52]{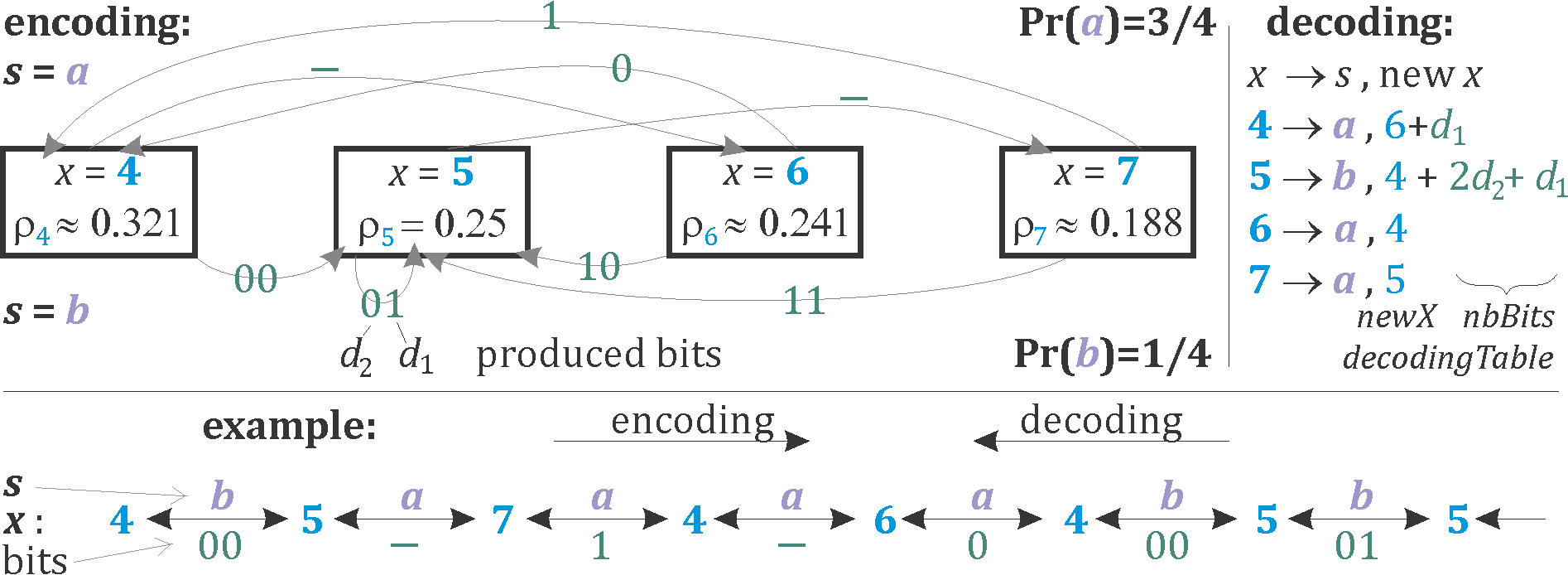}
\end{center}
\caption{State machine for encoding and decoding of ANS \cite{duda2014asymmetric} ($x$: state number)}\label{ans state machine}
\end{figure}

According to Shannon's source coding theorem, a flow of data can be compressed without information loss up to a certain bound determined by its entropy:
\hfill\break\hfill\break
\textbf{Source coding theorem \cite{entropy_bound}:}
\hfill\break\indent
\textit{$N$ drawings of a random variable $X$ with entropy $H(X)$ can be encoded using $NH(X)$ bits with a negligible probability of information loss as $N \xrightarrow{} \infty$. Conversely, they cannot be encoded using less than $NH(X)$ bits without the event of a loss of information being almost certain.}
\hfill\break\hfill\break\indent
Entropy coding assigns variable length codes to symbols according to their occurrence probabilities. The more frequent a symbol, the shorter its code, thus allowing to compress the overall dataflow. Shannon's source coding theorem yields that entropy coding is most useful when the distribution has low entropy. Huffman coding \cite{huffman_code} and arithmetic coding \cite{arithmetic_code} are the two most famous types of entropy coding algorithms. Huffman coding requires at least 1 bit per symbol, which reduces its compression capacity. Arithmetic coding is complex to implement due to its reparametrization.
\hfill\break\indent
Asymmetric Numeral Systems (ANS) is a type of entropy coding introduced in \cite{duda2014asymmetric}. It comes in several variants, including the tabled ANS (tANS) variant, whose decoder is a finite automaton (Figure \ref{ans state machine}). It has the same asymptotic compression capabilities as arithmetic coding. We aim to use tANS in order to decode data as it transits from the DRAM to the computation unit of the chip.

The number of states $l$ for the automaton can be chosen arbitrarily. As shown on Figure \ref{ans state machine} in the case of a binary alphabet \{a,b\}, the encoding of a symbol is achieved by following a state transition in the automaton and concatenating the binary string that labels the transition to the encoding stream. Decoding is performed by running through the automaton in reverse order according to the bits read in the encoded stream. One may notice that some transitions have empty labels: this is where tANS comes into its own compared with Huffman coding. Indeed, Huffman encodes each symbol using at least one bit, disallowing the crossing of the 1-bit frontier. ANS makes this crossing possible thanks to the unlabeled transitions (the -- signs in Figure \ref{ans state machine}), which occur when no bit needs to be read from the bitstream to decode a symbol.
\hfill\break\indent
\cite{duda2014asymmetric} shows that tANS compression rate depends on the initialization of the states. The more states are available in the state machine, the closer tANS comes to Shannon's entropy bound. The gap $\Delta H$ between the code length and the entropy of the distribution decreases quadratically with the number of states as in Equation \ref{eqn:deltaHbound}, with $p_s$ the occurrence probability of symbol $s$:

\begin{equation} \label{eqn:deltaHbound}
    \Delta H \leq \frac{1}{l^2 ln(4)} \sum_s \frac{1}{p_s}(\frac{p_s}{2\text{min}_{s'}p_{s'}} + \frac{1}{2})^2 + \mathcal{O}(l^{-3}) 
\end{equation}

\begin{figure}[h]

\begin{center}
\includegraphics[scale=0.4]{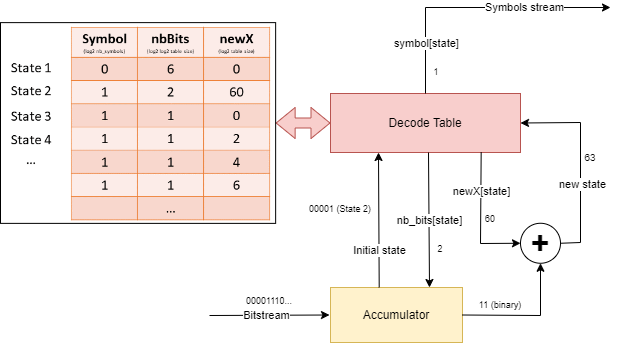}
\end{center}
\caption{ANS decoder diagram.}\label{ANS decoder}
\end{figure}

We propose an implementation of the ANS decoder using a Look-Up Table (LUT). The stream of compressed weights enters a First-In First-Out buffer that interacts with the decode table. The initial state is sent to the table, which outputs the decoded symbol, a number of bits $nb\_bits$ to read from the buffer and a transition integer $newX$. 
\hfill\break\indent
All of the LUT information can be stored using $3l$ bytes (one symbol, one number of bits and one integer per state, which can all be stored on one byte). Figure \ref{ANS decoder} represents a 64-states decoder with a 6-bit adder for 4-bit symbols. In this example, the whole LUT can be stored over 192 bytes. 

This implementation makes tANS compatible with a dataflow model, and does not require context information to decode (contrary to \cite{deepcabac} for instance). Next, reducing decoding latency requires to make tANS compliant with the Single-Instruction-Multiple-Data (SIMD) paradigm of computation (parallel computing). Therefore, we propose to split the dataflow of weights across channels and to reuse the same decoder over the concurrent streams. This will be more thoroughly covered in section \ref{parallelization}.

\subsection{Zero-point quantization} \label{zero-point quantization}

Uniform quantizers are among the easiest to implement types of quantizers, and have been successfully leveraged in state-of-the-art quantization methods \cite{esser2020learned,jin2019efficient,fracbits}. Usually, weights are quantized over some number of bits. However, one could very well use any number of quantization bins, which does not need to be a power of two. Specifically, we choose a quantizer with a zero-point (and therefore an odd number of quantization points) that keeps a symetric quantization points distribution, to match the typical symetric gaussian shaped distribution of weights of full precision neural networks. Indeed, the natural distribution of weights of a trained neural network tends to concentrate around 0, which means that the zero-point of our quantizer would capture an important part of the distribution, thus contributing to reduce the entropy of quantized weights and increase the efficiency of our coding scheme. In order to challenge this intuition, we train a Resnet-18 \cite{resnet} (16-32-64-128) on the dataset CIFAR-10 \cite{cifar} for various numbers of quantization bins and compare the results in terms of entropy ($H \times |W|$, in MB) and accuracy (Top-1 Acc, in \%). We quantize its weights using SAT \cite{jin2019efficient}. Our findings are reported in Table \ref{objectif vs quantif}.

\begin{table}[h]
\caption{Network Top-1 Accuracy on CIFAR-10 and Shannon compression bound using various quantization precisions.}\label{objectif vs quantif}
\begin{center}
\begin{tabular}{ccccc} \toprule
    {Network} & {Quantization} & {0-point} & & \multicolumn{1}{c}{\textbf{MB/Top-1}} \\ \midrule
    {\multirow{8}{*}{\parbox{1.8cm}{\centering Resnet-18}}} 
    & {\multirow{2}{*}{4-bit}} & \multirow{2}{*}{No} & {\multirow{2}{*}{\parbox{1.3cm}{\centering $H\times |W|$ \\ Top-1 Acc}}} & 0.23\\
    & & & & 90.63 \\ \cmidrule{2-5}
    & {\multirow{2}{*}{3-bit}} & \multirow{2}{*}{No} & {\multirow{2}{*}{\parbox{1.3cm}{\centering $H\times |W|$ \\ Top-1 Acc}}} & 0.13 \\
    & & & & 90.71 \\ \cmidrule{2-5}
    & {\multirow{2}{*}{5 bins}} & \multirow{2}{*}{Yes} & {\multirow{2}{*}{\parbox{1.3cm}{\centering $H\times |W|$ \\ Top-1 Acc}}} & \textbf{0.07} \\
    & & & & 90.21  \\ \cmidrule{2-5}
    & {\multirow{2}{*}{2-bit}} & \multirow{2}{*}{No} & {\multirow{2}{*}{\parbox{1.3cm}{\centering $H\times |W|$ \\ Top-1 Acc}}} & 0.09\\
    & & & & 89.63 \\
    \bottomrule
\end{tabular}
\end{center}
\end{table}

As can be seen in the previous table, the 5 bins quantization scheme (with zero-point) exhibits the lowest entropy bound and remains within range of integer-bit quantization precisions in terms of Top-1 Accuracy. A lower precision reduces the entropy, but it turns out that the zero-point helps reducing it even more and with less impact on network accuracy (5 bins has a lower entropy bound than 2-bit quantization and reaches better accuracy by 0.6\%). Table \ref{objectif vs quantif} also shows that there is a strong interplay between the quantization precision and the entropy bound. This observation justifies the use of a mixed-precision quantization method with a zero-point to reach the lowest quantization precision in each particular layer while preserving task performance as much as possible.

\subsection{Training}

The objective of entropy reduction should guide the choice of the quantization precisions in our mixed precision scheme. Fracbits \cite{fracbits} integrates the quantization precision selection directly into the learning process by learning a floating-point bit precision parameter $\lambda$ and then rounding it. Fracbits uses two different criteria, depending if activation quantization is enabled or not. The weight-only criterion ($\mathcal{L}^{size}$) on the model size is described in Equation \ref{eqn:fracbits_size}:

\begin{equation} \label{eqn:fracbits_size}
    \mathcal{L}^{size} = |\sum_w \lambda_w - \widehat{\mathcal{L}}|
\end{equation}

where $\lambda_w$ is the precision parameter attached to weight $w$ (whether it is layerwise or channelwise quantization) and $\widehat{\mathcal{L}}$ is the compression goal. We adapt this criterion (and turn it into $\mathcal{L}^{entropy}$) so that the precisions are driven by the entropy bound rather than by the raw memory footprint (Equation \ref{eqn:fracbits_entropy}):
\begin{equation}
\begin{aligned} \label{eqn:fracbits_entropy}
    \mathcal{L}^{entropy} = \frac{1}{\hat{H}} (|\sum_W |W| ((1 - \lambda_W + \lfloor \lambda_W\rfloor) H_{W(\lfloor \lambda_W\rfloor) }\\ + (\lambda_W - \lfloor \lambda_W\rfloor) H_{W(\lceil \lambda_W \rceil)}) - \hat{H}|)
\end{aligned}
\end{equation}
where $W$ are the parameter groups of the network, 
$\lambda_W$ the Fracbits precision parameters of these groups, $H_{W(k)}$ the entropy of $k$-bins quantized weights, and $\hat{H}$ the total entropy goal. The gradient $\frac{\partial \mathcal{L}^{entropy}}{\partial \lambda_W}$ is computed following Equation \ref{eqn:fracbits_entropy_diff}:
\begin{equation} \label{eqn:fracbits_entropy_diff}
\frac{\partial \mathcal{L}^{entropy}}{\partial \lambda_W} = \frac{|W|}{\hat{H}}(H_{W(\lceil \lambda_W \rceil)} - H_{W(\lfloor \lambda_W\rfloor) })
\end{equation}
The precision parameters $\lambda_W$ can thus be learned with our entropy goal. Lastly, in order to take advantage of the findings of \ref{zero-point quantization}, we reinterpret the meaning of the $\lambda_W$ parameters: instead of $\lambda_W$ being a number of bits, $\lfloor 2 \times \lambda_W + 1 \rceil$ will be an odd number of quantization bins (so that the middle one is a zero-point bin).

\section{Experiments}

\subsection{ImageNet results}

Our method was evaluated on the ImageNet classification task \cite{russakovsky2015imagenet}. All experiments were run using PyTorch \cite{pytorch}. We train Resnets \cite{resnet} using our modified Fracbits training pipeline and compress the weights of each layer using a 256 states tANS coder. The first and last layers remain in 8-bit and remain uncompressed. In Table \ref{table:imgnet_compared}, we report the total memory cost of the network storage and the entropy bound.

\begin{table}[h]
\caption{Top-1 Accuracy and total memory cost of trained networks, ImageNet, state-of-the-art network compression methods for comparison.} \label{table:imgnet_compared}
\begin{center}
\begin{tabular}{ccccc} \toprule
    {Network} & {Method} & {Top-1 Acc} & {Size} & {$H \times |W|$} \\ \midrule
    {\multirow{5}{*}{\parbox{2cm} {\centering Resnet-18\\ 69.8\% \\ 46.8 MB}}} & HEMP & 68.8\% & 3.6 MB \\ \cmidrule{2-5}
    &  Oktay et al. & 70.0\% & 2.78 MB \\ \cmidrule{2-5}
    &  Fracbits & 68.5\% &  2.01 MB &  1.99 MB \\ \cmidrule{2-5}
    & Ours & 69.0\% &  \textbf{1.67 MB} &  1.65 MB \\  \midrule
    {\multirow{6}[4]{*}{\parbox{2cm}{\centering Resnet-50\\ 76.1\% \\  102.5 MB}}} & HEMP & 71.3\% & 5.5 MB \\ \cmidrule{2-5}
    & Oktay et al. & 73.5\% & 5.91 MB \\ \cmidrule{2-5}
    & DCABAC & 73.7\% & 5.25 MB \\ \cmidrule{2-5}
    & DCABAC & 74.5\% & 10.4 MB \\ \cmidrule{2-5}
    & Ours & 75.6\% &  \textbf{4.52 MB} & 4.39 MB \\  \bottomrule
\end{tabular}
\end{center}
\end{table}

These results show at least comparable Top-1 Accuracy and better compression rate for our zero-point + mixed precision quantization approach over the reported state-of-the art methods \cite{oktay2020scalable,hemp,deepcabac}. Furthermore, the decoders of these compared methods are much more complex than ANS and practically impossible to use at inference time. The accuracy drop compared to full-precision networks is below $1\%$ with a higher than $20 \times$ compression ratio.
\hfill\break\indent
To confirm the actual role of the zero-point in lowering the entropy bound, we run an experiment with standard Fracbits on Resnet-18 (Fracbits) and a 2MB goal. It achieves $68.5 \%$ Top-1 Accuracy, worse than with our approach, and with a worse compression rate. This result proves the interest of the zero-point for memory reduction.
\hfill\break\indent
The following section addresses the hardware feasibility of ANS by simulating the parallelization of the decoder and by evaluating the impact of the decoding table size reduction.

\subsection{Table size and decoding parallelization} \label{parallelization}

The decoder needs to be compatible with the SIMD paradigm of computation in order to be usable at inference time. The same decoder is used to decode multiple streams at a time following Figure \ref{parallel decode}. Instead of having one single state, the parallel ANS decoder has a vector of states (one state for each stream). The latency of the parallel decoder depends on the parallelism degree. If weights are encoded in parallel streams of $n$ weights each, decoding amounts to $n$ LUT accesses + $n$ low bit additions per stream.

\begin{figure}[H]

\begin{center}
\includegraphics[scale=0.45]{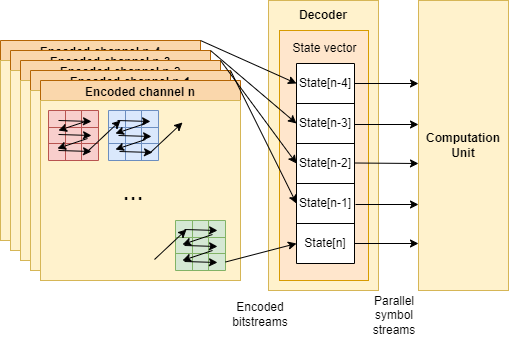}
\end{center}
\caption{Decoder parallelization scheme.}\label{parallel decode}
\end{figure}

Using the same Resnet-50 as in Table \ref{table:imgnet_compared}, weights are grouped by input channel and separately encoded. Table \ref{parallel_decoding_table} reports the total memory footprint of the network depending on whether the decoding process is parallelized or not and on the size of the decoding table, which can be chosen.

\begin{table}[h] 
\caption{Impact of parallelization and table size reduction on final compressed network size.} \label{parallel_decoding_table}
\centering
\begin{tabular}{cccc} \toprule & & \multicolumn{2}{c}{\textbf{Compressed size}} \\
    {Network} & {Table size ($l$)} & {Single stream} & {Parallel streams}\\ \midrule
    {\multirow{3}[4]{*}{Resnet-50}} & 256 & 4.52 MB & 4.54 MB\\ \cmidrule{2-4}
    & 128 & 4.68 MB & 4.70 MB\\ \cmidrule{2-4}
    & 64 & 5.02 MB & 5.04 MB\\
    \bottomrule
\end{tabular}
\end{table}

Parallelization hardly increases the network's memory footprint, with less than $1\%$ overhead. As for the table size, $256$ states are necessary to reach a sub-$3\%$ relative gap to the entropy bound ($4.39$ MB). However, it is possible to use only 64 states with less than $15\%$ relative gap.

\subsection{Analysis}

\begin{figure}[H]

\begin{center}
\includegraphics[scale=0.26]{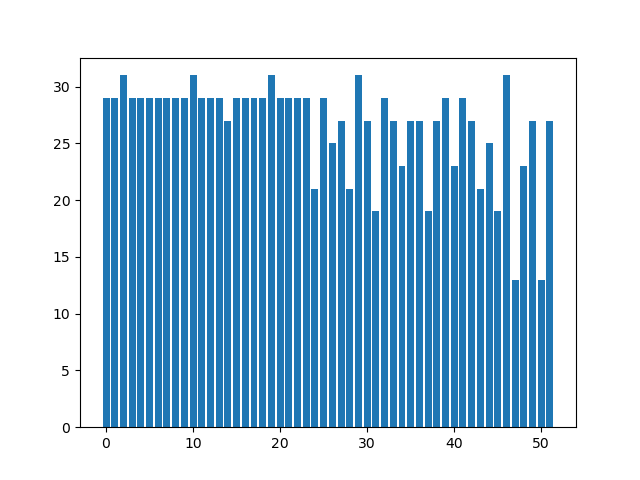}
\end{center}
\caption{Distribution of quantization precision among layers, Resnet-50, number of bins.}\label{precision per layer}
\end{figure}

\begin{figure}[H]
\begin{center}
\includegraphics[scale=0.26]{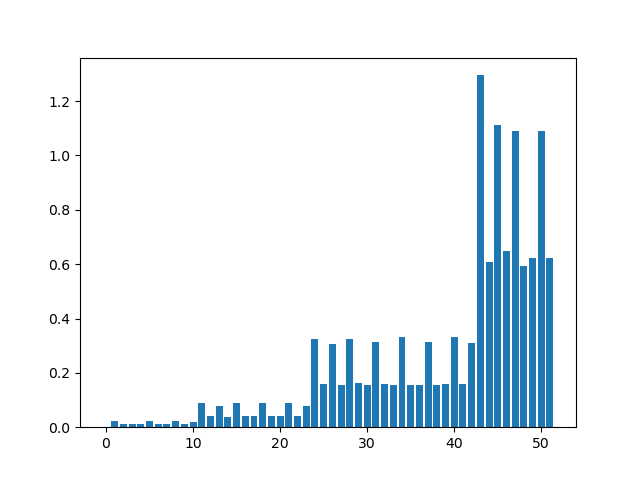}
\includegraphics[scale=0.26]{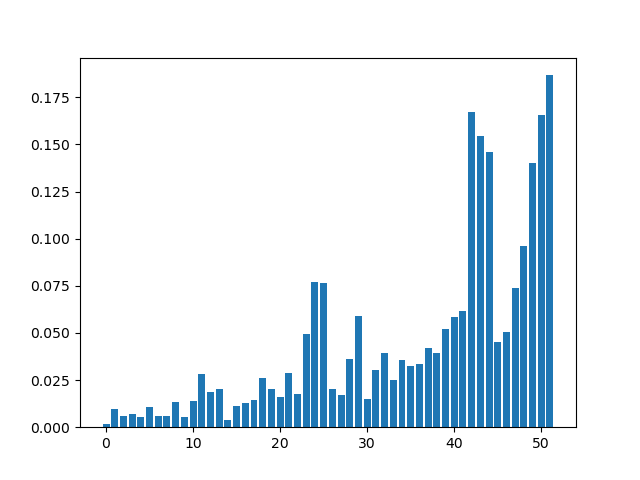}
\end{center}
\caption{Memory cost distribution (MB, vertical axis) among Resnet-50 layers. Quantized uncompressed network (up), and quantized ANS compressed (down). Axes scales are different.}
\label{entropy by layer}
\end{figure}

In this section, we analyze the distribution of memory complexity within the layers of the network. First, Figure \ref{precision per layer} shows the number of quantization bins allocated to each layer at the end of the training. The network is initialized with $31$ bins ($<$ 5-bit), and the first half of the network remains at this precision. At the end of the network, the precisions of the most abundant layers have been reduced by the training (down to $13$ bins for two of them). 
Figure \ref{entropy by layer} represents the layerwise memory cost. Some layers are compressed up to $20 \times$, and the peak bandwidth decreases from 1.3 MB to 185 KB thanks to ANS coding, which amounts to an $86\%$ decrease. The total storage cost for the feature extractor of this network is $2.5$ MB ($4.5$ in total, $2.0$ for the uncompressed 8-bit classifier alone). The average compression rate reaches $0.85$ bit per weight in these layers, meaning they are more compressed than if fully binarized.

\section{Conclusion and future works}

Reducing memory bandwidth and overall memory footprint of neural networks is a major challenge in deploying models on chip, translating to smaller and less resource-consuming chips. In this paper, we have shown that combining integer quantization with zero-point and entropy decoding is a viable way of drastically reducing the model footprint while doing a much better job at preserving its accuracy than quantization-only approaches, with less than $1\%$ accuracy drop and sub-binary compression rates reached over Resnets in the ImageNet classification task. We also identified one strong candidate for an inference-time entropy decoder.
\hfill\break\indent
Future works include the improvement of the quantization-aware training pipeline aiming at actively reducing the entropy of the parameters, activation quantization for faster inference, and the hardware implementation of ANS decoding.

\section*{Acknowledgment}

This work was supported by a French government grant managed by the Agence Nationale de la Recherche under the France 2030 program with the reference "ANR-23-DEGR-0001".

\newpage

\bibliographystyle{plain} 
\bibliography{conference_101719}

\end{document}